\pdfoutput=1

\documentclass[11pt]{article}

\usepackage[]{EMNLP2023}

\usepackage{times}
\usepackage{latexsym}

\usepackage[T1]{fontenc}

\usepackage[utf8]{inputenc}

\usepackage{microtype}

\usepackage{inconsolata}

\usepackage{booktabs}
\usepackage{multirow}
\usepackage{array}
\usepackage{graphicx}
\usepackage{float}
\usepackage{amsmath, amsfonts, amssymb}
\usepackage{stfloats}
\usepackage{CJKutf8}
\usepackage{colortbl}
\usepackage{amssymb}

\newcolumntype{P}[1]{>{\centering\arraybackslash}p{#1}}

\usepackage{cleveref}
\crefname{section}{\S}{\S\S}
\crefname{table}{Table}{}
\crefname{figure}{Figure}{}
\crefname{appendix}{Appendix}{}

\title{Separating form and meaning: Using self-consistency to quantify task understanding across multiple senses}

\newcommand{\osnabruek}{$^{\Omega}$}

\author{
    Xenia Ohmer\osnabruek \and Elia Bruni\osnabruek\footnotemark[1] \and Dieuwke Hupkes$^{\infty}\footnotemark[1]$\\
\osnabruek Osnabrück University \hspace{5mm} $^\infty$ FAIR \\
\texttt{\{xenia.ohmer, elia.bruni\}@uni-osnabrueck.de}\\
\texttt{dieuwkehupkes@meta.com}
}

\definecolor{ocean}{RGB}{30,150,180}

\begin{document}
\maketitle

\setcounter{footnote}{0}

\begin{abstract}
At the staggering pace with which the capabilities of large language models (LLMs) are increasing, creating future-proof evaluation sets to assess their understanding becomes more and more challenging.
In this paper, we propose a novel paradigm for evaluating LLMs which leverages the idea that correct world understanding should be consistent across different (Fregean) senses of the same meaning. 
Accordingly, we measure understanding not in terms of correctness but by evaluating consistency across multiple senses that are generated by the model itself.
We showcase our approach by instantiating a test where the different senses are different languages, hence using multilingual self-consistency as a litmus test for the model's understanding and simultaneously addressing the important topic of multilinguality.
Taking one of the latest versions of ChatGPT as our object of study, we evaluate multilingual consistency for two different tasks across three different languages.
We show that its multilingual consistency is still lacking, and that its task and world understanding are thus not language-independent.
As our approach does not require any static evaluation corpora in languages other than English, it can easily and cheaply be extended to different languages and tasks and could become an integral part of future benchmarking efforts. 
\end{abstract}

\section{Introduction}

The staggering pace at which the capabilities of large language models (LLMs) have increased in the recent past comes with many questions related to what kind of progress we are making on the road towards true machine intelligence and human-level understanding.
To assess such progress, practitioners often rely on benchmarks that measure natural language understanding \citep[e.g.][]{williams-etal-2018-broad,nie-etal-2020-adversarial}, commonsense reasoning \citep[e.g.][]{sap-etal-2019-social, Bisk2020}, or probe for factual knowledge \citep[e.g.][]{hendrycks2021measuring}, among other things.
The extent to which such benchmarks can be used to assess whether LLMs ``understand'' language is widely debated \citep[e.g.][]{mitchell_krakauer_2023,raji2021everything}.
Often mentioned concerns in this context are that LLMs may learn specific lexical patterns rather than general principles \citep[e.g.][]{ray-choudhury-etal-2022-machine} and, relatedly, that benchmark scores may confuse competence in \emph{form} with competence in \emph{meaning} \citep[e.g.][]{heinemanrethinking}.
In support of these concerns, LLMs have been found to bypass certain tasks by relying on memorised information from the training data \citep{mckenna2023sources}.
More recently, the enormous amount of data that models are trained on and the fact that this data is often not publically accessible have further increased the difficulty of assessing whether benchmarks really quantify what they are meant to quantify.
A benchmark always makes assumptions about what a model has seen in its training phase, and, given the rapid changes on that front, it is difficult to design challenging benchmarks that remain informative past training rounds of new models.
In addition, novel evaluation data may leak into the training data of newly trained models\footnote{E.g.\ portions of the BIG-Bench data \citep{srivastava2022beyond} were inadvertently added to the GPT4 training corpus \citep[footnote 5]{openai2023gpt4}.} -- which even the most future-proofed benchmarks may not withstand.

In this paper, we propose a novel approach to evaluate models' task or world understanding that aims to create some separation between form and meaning in benchmarking and simultaneously mitigates the challenging evaluation-contamination loop.
Our method is based on the idea that language is used to describe or act in the world \citep{wittgenstein2010philosophical} and that this world functions as an anchor for diverse linguistic forms.
Having a genuine understanding of the world thus implies consistency among different linguistic expressions that pertain to the same entities within the world.
To give an example, if you ask a colleague who is fluent in both French and English if a particular statement is true, you expect their answer to be invariant to the language (French or English) in which you ask this question.
We leverage this intuition to investigate whether models have a consistent world model across different senses (in the case above: languages) and, consequently, a consistent understanding of the tasks that they are asked to execute.
Loosely inspired by \citet{frege_1892}, we take different senses to be different modes of presentation or notational variants.
Crucially, rather than generating different senses ourselves, we ask the model to create different versions of the same question.
This ensures that potential inconsistencies are really due to model-internal inconsistencies rather than misinterpretations of the question.
Additionally, the method is protected from data contamination: as the different senses are regenerated for every evaluation, they cannot leak into new training data.
Lastly, it can cheaply and easily be applied to already available benchmarks and therefore reduces the burden on data generation.

\begin{figure*}[ht!]
    \centering
    \includegraphics[width=\textwidth]{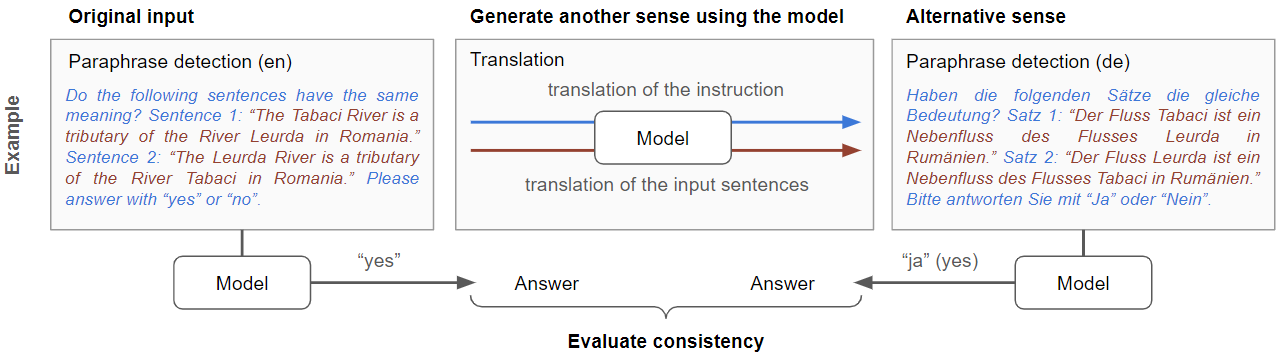}
    \caption{Illustration of the basic mechanism of our paradigm: We use the model to generate other senses of the original input.
    The model's answers on the original input and the alternative sense are used to evaluate its consistency.
    In this example, the model is presented with the task of paraphrase detection in English (sentences taken from PAWS-X) and generates another sense by translating from English to German.}
    \label{fig:paradigm}
\end{figure*}

Our approach can be applied to a number of different senses.
Here, we showcase it focusing on the multilingual case described in the example above, by asking whether models are consistent across different languages (see \cref{fig:paradigm}).
In essence, we are thus using multilingual self-consistency as a litmus test for their understanding, simultaneously addressing the important topic of multilinguality.
Taking one of the latest SOTA versions of ChatGPT\footnote{\url{https://openai.com/blog/chatgpt}} as our object of study, we evaluate multilingual understanding for two different tasks (paraphrase detection and natural language inference) across three different languages (Chinese, German and English).
It turns out that the model is inconsistent across all languages and tasks, despite being able to perform the tasks reasonably well in English and generating high-quality translations.
Taken together, our analyses provide strong evidence that the model's task understanding is modulated by the representational form of the task.

In sum, we make the following contributions:\vspace{-1mm} \begin{itemize}\setlength\itemsep{0.5mm}
\item[i)] We introduce multisense consistency as a novel, cheap, and data-contamination-proof evaluation paradigm for LLMs;
\item[ii)] We showcase this paradigm by implementing a specific version that utilises \emph{multilinguality} to create different senses;
\item[iii)] Using this implementation, we evaluate ChatGPT to illustrate that multilingual consistency of SOTA LLMs is still lacking;
\item[iv)] With a range of ablation experiments (see \cref{fig:experiments_overview}), we demonstrate that the observed inconsistencies in fact arise from a language-dependent task understanding (rather than an inability to translate or perform the task).
\end{itemize}

With our work, we hope to not only present an interesting set of empirical results on multilingual consistency but also propose a novel, easily applicable method to generate many more challenging evaluation tests.
Our framework targets models that can follow instructions to generate alternative senses and are able to generate these senses based on these instructions.
Thus, with the growing popularity and capabilities of chat-models and instruction-tuned models, such as GPT-4 \citep{openai2023gpt4} or Llama-2 \citep{touvron2023llama}, our framework is becoming increasingly relevant.\footnote{Our code is available at \url{https://github.com/XeniaOhmer/multisense_consistency}.}

\section{Related work}

Existing benchmarks for \textit{evaluating language understanding} in LLMs form the foundation for our work.
The main idea of our paradigm is to evaluate LLMs in terms of their consistency across different senses of these benchmarks and is therefore related to other work on \textit{self-consistency in LLMs}.
In creating multiple senses through translation, there is also a close connection between our execution of this paradigm and \textit{multilingual evaluation}.
\cref{app:genbench} provides a GenBench eval card \citep{hupkes2023stateoftheart} that classifies our work in the context of generalisation research.

\paragraph{Evaluating language understanding.}
A wide range of benchmark tasks has been developed to evaluate specific aspects of natural language understanding in LLMs.
To evaluate \textit{general} language understanding across diverse tasks, genres, and datasets, several of these tasks have been combined into multi-task benchmarks, such as GLUE \cite{wang-etal-2018-glue} or SuperGLUE \cite{wang-2019-etal-superglue}.
These benchmarks focus on English and evaluate, among others, paraphrase identification \cite[e.g.\ PAWS; ][]{zhang-etal-2019-paws}, natural language inference \cite[e.g.\ MNLI; ][]{williams-etal-2018-broad}, and commonsense reasoning \cite[e.g.\ COPA; ][]{copa_roemmele}.
In response to the rapid improvements of LLMs on these benchmarks other multitask benchmarks have been developed.
MMLU, for example, assesses world knowledge and problem-solving ability across a large number of subjects, covering STEM, humanities, social sciences, and more \cite{hendrycks2021measuring}.
While our paradigm also makes an effort to find more appropriate evaluation methods, it not only assesses performance but also evaluates the model's ability to consistently solve a task across multiple languages, thereby providing insights into its ability to abstract from specific representational forms.

\paragraph{Self-consistency in LLMs.}
Various studies have shown that inconsistencies are common in LLMs (and suggested methods for improving consistency, which is not our focus). 
These studies are mostly concerned with self-consistency in natural language inference (NLI) \cite[e.g.][]{minervini-riedel-2018-adversarially, wang2018simply, li-etal-2019-logic, hosseini-etal-2021-understanding} and question answering \cite[e.g.][]{kassner-schutze-2020-negated, alberti-etal-2019-synthetic, mitchell-etal-2022-enhancing, chen-etal-2021-nli-models, elazar-2021, kassner-etal-2021-beliefbank, asai-hajishirzi-2020-logic, hosseini-etal-2021-understanding}. 
For example, \citet{kassner-etal-2021-beliefbank} created a dataset to measure a model's consistency by 
evaluating its responses to sentence pairs that are subject to certain constraints (e.g.\ if \textit{X is a dog} is true, \textit{X has a tail} must also be true). 
More similar to our work, \citet{elazar-2021} studied whether factual knowledge in masked language models is invariant to paraphrasing.
To this end, they created \textsc{ParaRel}, a dataset containing cloze-style English paraphrases.
In these two examples, consistency is either evaluated against a network of logical relationships between beliefs or by generating different forms of the same meaning through paraphrasing.
BECEL \citep{jang-etal-2022-becel} is a benchmark for evaluating these two types of consistency (logical and semantic) across various tasks.
This benchmark has recently been used to evaluate ChatGPT, showing that it is more consistent for negations than other LLMs, but still likely to generate different responses to paraphrases of the same meaning \citep{jang2023consistency}.
Unlike previous work -- except \citep{jang2023consistency} -- we focus on true \emph{self}-consistency: Different forms of the same meaning are generated by the model itself, rather than externally.

\paragraph{Multilingual evaluation.} The development of cross- and multi-lingual LLMs has spurred interest in multilingual evaluation beyond translation.
Several multilingual versions of benchmark tasks have been generated, such as PAWS-X \cite{yang-etal-2019-paws}, XCOPA \cite{ponti-etal-2020-xcopa}, and XNLI \cite{conneau-etal-2018-xnli} -- usually through expert translations from the original task \cite[for a more expansive overview, we refer to][Appendix D]{hupkes2023stateoftheart}.
In addition, multilingual tasks have been combined to form multilingual multitask benchmarks, including XTREME \cite{hu-etal-2020-xtreme}, XTREME-R \cite{ruder-etal-2021-xtreme}, and XGLUE \cite{liang-etal-2020-xglue}.
All of these benchmarks reveal language-dependent differences in performance for current multilingual LLMs.
Our approach is different in that we aim to evaluate self-consistency by detecting language-dependent changes in model responses, relying on the model's own translations instead of external translations.

\section{Methods}\label{sec:methods}

We now proceed with describing the model (\cref{subsec:model}) and the benchmark data (\cref{subsec:benchmarks}) we use for our experiments, as well as the procedure we use for extracting translations from the model (\cref{subsec:translations}).

\subsection{Model and hyperparameters}\label{subsec:model}

We showcase our paradigm using  \textsc{gpt-3.5-turbo-0301}.
We use the default parameters but set the temperature to $0.25$.
We found a low temperature to yield model responses that more closely match the template answers for benchmarking, as well as model translations that better capture the meaning of the source sentences.
In addition, we set the maximal number of generated tokens to $256$ for benchmarking and $2048$ for translation.

\subsection{Benchmarking}\label{subsec:benchmarks}

\paragraph{Tasks and languages.} 
We evaluate understanding using the multilingual benchmarks PAWS-X and XNLI (test splits).
While our paradigm does not require parallel multilingual datasets, we use them here to evaluate translation quality, compare translations between two languages in both directions, and analyse differences that arise from using model-internal instead of model-external translations.
PAWS-X is an adversarial paraphrase identification task, consisting of sentence pairs created by word-swapping, resulting in negative pairs that have clearly distinct meanings, but a high lexical overlap (see, for instance, the example in \cref{fig:paradigm}).
XNLI, on the other hand, is an NLI benchmark, containing sentence pairs where one sentence entails the other, contradicts it, or neither of the two (neutral).
Importantly, on either task, the model's judgment should not be dependent on nuances in meaning that may be lost in translation.
For our experiments, we focus on the German, English, and Chinese partitions of the respective benchmarks.

\paragraph{Instructions.} 
We design task instructions in English to evaluate the model's zero-shot performance.
Given that the benchmarks are binary/ternary classification problems, the instructions can be formulated such that the model's responses can be easily standardised and evaluated:
\vspace{-1mm}\begin{itemize} \setlength\itemsep{0.5mm}
    \item {PAWS-X: \textit{Do the following sentences have the same meaning? Sentence 1: ``[sentence\_1]'' Sentence 2: ``[sentence\_2]'' Please answer with ``yes'' or ``no''.}}
    \item {XNLI: \textit{Given the following sentence pair, which one of the following is true: (A) the first sentence entails the second sentence, (B) the first sentence contradicts the second sentence, or (C) neither of the two? Sentence 1: ``[sentence\_1]'' Sentence 2: ``[sentence\_2]'' Please answer with ``A'', ``B'', or ``C''.}}
\end{itemize}
In addition, these instructions are translated into German and Chinese by native speakers (see \cref{app:instructions}), in sum giving us ground truth input data and instructions in each language.

\paragraph{Evaluation.}
We process each input in a separate request.
We only accept model responses matching the template answer (e.g.\ ``yes'') or containing it (e.g.\ ``Yes, the sentences have the same meaning.''), ignoring casing.\footnote{Only a negligible amount of responses do not fall into one of these two categories ($<1\%$). These are mapped onto an additional label indicating invalidity.}
In the second case, we apply a semi-automatic standardisation procedure: a function maps the model's responses to one of the template answers, and these mappings are checked, and if necessary corrected, by hand.
Using the standardised responses, we can calculate the model's accuracy on the task, as well as the model's consistency across different runs.

\subsection{Model-internal translations}\label{subsec:translations}

We experiment with translations from English to Chinese and German, and from Chinese and German to English.
The original English, Chinese, and German tasks serve as baselines for our simulations.
Our main goal is to evaluate the consistency between the model's responses on these baselines and the model's responses on a model-internally generated translation, always comparing the source language to the translation from source to target.

\paragraph{Translation procedure and notation.}
We generate model-internal (zero-shot) translations of both input data and instructions.
The translation instructions (see \cref{app:translation_instructions}), written by native speakers, are always given in the source language.
For the task instructions, the model translates the instruction prefix (e.g.\ \textit{Do the following sentences have the same meaning?}), the word for sentence (e.g.\ \textit{sentence}), and the instruction suffix (e.g.\ \textit{Please answer with ``yes'' or ``no''.}) in separate requests, and we recompose these translations to an instruction in the target language (see \cref{app:instructions}).
For the input data, the model translates each sentence per input sentence pair in a separate request.

In what follows, we will denote the instruction of a task $T$ with $I$ and the input to which it is applied with $X$.
We annotate the language in which either of those is given with a subscript, which also indicates whether it is a model translation from another language.
Thus, $T_{en}$ refers to the scenario in which both the instruction and the input are given in English, using the original benchmark data, while $T_{en\rightarrow de}$ denotes the model's translation of instruction and input sentences from English to German.
Following the same principle, $I_{en\rightarrow de}$ and $X_{en\rightarrow de}$ indicate instructions and input, respectively, that the model has translated from English to German.

\paragraph{Evaluation.} The model's translations of the task instructions were reviewed by native speakers who found the translations to be appropriate, apart from slight deviations in the translations from Chinese to English: For PAWS-X the instructions mention a single sentence instead of a sentence pair (\textit{Does the following sentence have the same meaning?}) and for XNLI the word \textit{covers} is used for \textit{entails}.
To evaluate the quality of the model's translations of the actual input sentences, we employ BLEU scores \cite{papineni-etal-2002-bleu} calculated with SacreBLEU \cite{post-2018-call}, as well as ROUGE \cite{lin-2004-rouge}, and COMET-22 \cite{rei-etal-2022-comet} scores (see \cref{subsec:translation_evaluation}).

\begin{figure*}
    \centering
    \includegraphics[width=0.85\textwidth]{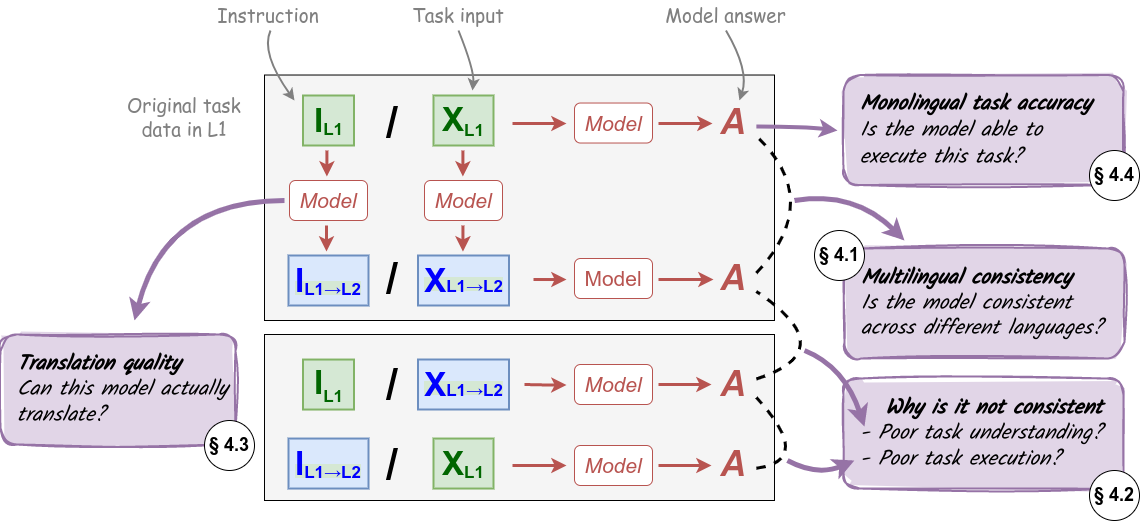}
    \caption{An overview of our experiments and analyses.}
    \label{fig:experiments_overview}
\end{figure*}

\section{Results}\label{sec:results}

In this section, we discuss the results of our experiments (see \cref{fig:experiments_overview}), beginning with our primary experiment in which we assess how consistent the model’s task
understanding is across languages (\cref{subsec:multilingual_consistency}).
In subsequent experiments, we investigate the individual effects of translating the dataset or the instructions (\cref{subsec:interpretation_execution}), and ensure that inconsistencies do not arise from inaccurate translations (\cref{subsec:translation_evaluation}) or poor task performance (\cref{subsec:consistency_performance}).

\subsection{Multilingual consistency}\label{subsec:multilingual_consistency}

In our primary experiment, we assess the consistency of a model's understanding by comparing the model's responses in a monolingual setting -- with the original input data and instruction language -- with its responses when using a model-internal translation of those.
Crucially, the task translations are produced by the model itself, rather than externally.
Assuming that the model is a good model of translation (see \cref{subsec:translation_evaluation}), its translations should be meaning-preserving.
In that case, \emph{if} the model has a meaning-based task understanding, its responses to both task versions should be consistent.

The results are reported in column \textit{T} (\emph{Task}) of \cref{tab:consistency}.
As we can see, there is not a single case where the scores are near-maximal, indicating that the task understanding of the model is not consistent across the evaluated languages.
Regarding the language pairs, consistencies tend to be higher when translating between English and German compared to English and Chinese, with an exception for translations \textit{to} English on XNLI (bottom rows).
More details on the differences in predictions before and after translation can be found in \cref{app:elaboration_main_experiment} and a qualitative analysis of the translations from English to German in \cref{app:qualitative_analysis}.
Besides, to provide an example where our paradigm is applied to a monolingual benchmark, we run the main experiment also for BoolQ \citep{clark-etal-2019-boolq}, which yields similar results (see \cref{app:boolq}).

\begin{table}[ht!]
\centering
\begin{tabular}{l|l|ccc}
 \multicolumn{2}{c|}{ } & \multicolumn{3}{c}{\textbf{Consistency}}\\
 & \textbf{Src}$\rightarrow$\textbf{Tgt} & $T$ & $I$& $X$\\
\hline
\multirow{4}{*}{PAWS-X}     & en$\rightarrow$de   &  0.84  & 0.93  &  0.85\\
                            & en$\rightarrow$zh   &  0.76  & 0.91  &  0.79\\
                            \cline{2-5}
                            & de$\rightarrow$en   &  0.86  & 0.93  &  0.86\\
                            & zh$\rightarrow$en   &  0.70  & 0.87  &  0.75\\
\hline
\multirow{4}{*}{XNLI}       & en$\rightarrow$de   &  0.74  & 0.81  &  0.76\\
                            & en$\rightarrow$zh   &  0.67  & 0.77  &  0.71\\
                            \cline{2-5}
                            & de$\rightarrow$en   &  0.63  & 0.69  &  0.81\\
                            & zh$\rightarrow$en   &  0.67  & 0.79  &  0.72\\
\end{tabular}
\caption{Consistency between baseline ($T_{src}$) and model-internal translation from source to target language ($T_{src\rightarrow tgt}$).
Shown are the consistencies for translating input data and instruction (column \textit{T}), instruction only (\textit{I}), or input data only (\textit{X}).}
\label{tab:consistency}
\end{table}

\subsection{Interpretation and execution consistency}\label{subsec:interpretation_execution}

When the model is inconsistent across languages, we need to determine whether it is due to an inadequate understanding of what it is asked to do in the target language or an inability to perform what it is asked to do in that language. 
We differentiate these effects by assessing the model's consistency when translating only the instruction, while retaining the original input sentences (e.g.\ comparing $T_{en}$ and $I_{en\rightarrow de} / X_{en}$) and its consistency when translating only the input sentences while preserving the original instruction (e.g.\ comparing $T_{en}$ and $I_{en} / X_{en\rightarrow de}$).
We show the results in \cref{tab:consistency}.

Neither consistencies for translating only the instructions (column \textit{I}) nor those for translating only the input sentences (column \textit{X}) are at their maximum, indicating that the model is inconsistent in both interpretation and execution.
Inconsistencies are consistently higher for PAWS-X than XNLI, probably because PAWS-X is a binary and XNLI is a ternary classification problem.
However, even translating a simple instruction, such as the one for PAWS-X, leads to inconsistencies for all translations.
Consistencies seem to decrease more when translating the input sentences compared to the instructions (except for German to English on XNLI) and even more when translating both (column \textit{T}).
Thus, inconsistencies in complete translations seem to be driven by differences in both task interpretation and execution, although differences in execution are more pronounced.

\subsection{Consistency and translation quality}\label{subsec:translation_evaluation}

The metric we propose in some way conflates monolingual task understanding and translation quality: inconsistencies can be driven by misalignment in task understanding, but also by poor translation quality.
While both are important, and the metric therefore favours models that do well across the board, it is worth further investigating which of the two drives the observed inconsistencies.

We start by considering the hypothesis that the model's consistency is suboptimal simply because it is not a good model for translation.
If the translation quality is poor, inconsistencies may arise from differences in meaning between original and translated inputs, rather than a language-dependent task understanding. 
To evaluate the model's translation quality specifically on the benchmark data, we examine the translations of the input data for all languages and directions using BLEU scores (see \cref{tab:bleu_scores}) and other commonly adapted metrics for translation quality (see \cref{app:translation_evaluation_scores}).

\begin{table}
\centering
\begin{tabular}{l|l|c}
 &   \textbf{Src}$\rightarrow$\textbf{Tgt} &  \textbf{BLEU} \\
\hline
\multirow{4}{*}{PAWS-X} &  en$\rightarrow$de    & 56.5 \\
                        & en$\rightarrow$zh     & 49.2 \\
                        & de$\rightarrow$en     & 60.0 \\
                        & zh$\rightarrow$en     & 37.6 \\
\hline
\multirow{4}{*}{XNLI}  &  en$\rightarrow$de     & 41.4 \\
                        & en$\rightarrow$zh     & 43.5 \\
                        & de$\rightarrow$en     & 45.8 \\
                        & zh$\rightarrow$en     & 28.0 \\
\end{tabular}
\caption{BLEU scores for the model-internal translation of the input data.}
\label{tab:bleu_scores}
\end{table}

All metrics indicate that the model's translations are of high quality across tasks and languages, with the sole exception of translations from Chinese to English.
The scores are generally higher for PAWS-X than XNLI, which might be due to the more challenging and diverse text sources used in generating XNLI.
The high scores thus suggest that, for most of the considered source-target language combinations, inconsistencies cannot be ascribed to changes in meaning induced by the translation.

\begin{table}[b]
\centering
\begin{tabular}{l|cccc}
& \multicolumn{4}{c}{$\boldsymbol{\rho}$ \textbf{(BLEU, consistency)}} \\
    & {\small en$\rightarrow$de} & {\small en$\rightarrow$zh} & {\small de$\rightarrow$en} & {\small zh$\rightarrow$en} \\
\hline
 PAWS-X  & 0.02 & 0.07 & 0.06 & 0.03 \\
 XNLI    & 0.03 & 0.02 & 0.05 & 0.09
\end{tabular}
\caption{Pearson correlation between BLEU scores and model consistency between original and translated inputs ($I_{source}$/$X_{source\rightarrow target}$).}
\label{tab:correlation_scores}
\end{table}

To further substantiate this claim, we compute the Pearson correlations between the BLEU score of the translation and the (binary) consistencies between the model's original responses and its responses on the translated benchmark data (see \cref{tab:correlation_scores}, top row for each task).
We focus on the simulations with model-internal translations of the input sentences, keeping the instruction in the source language (e.g.\ $I_{en} / X_{en\rightarrow de}$).
For these simulations, we can obtain a translation quality score per data point, which is not confounded with the translation quality of the instruction.
The BLEU score for a given data point is calculated by averaging the scores of the two sentences from the sentence pair.
All correlations are positive, yet, the absolute values are very low ($\le 0.09$).
These findings suggest that the observed inconsistencies are largely independent of the translation quality, at least in light of the generally high translation quality observed for this specific model.
Additional evidence is presented in \cref{app:translation_quality-consistency}, revealing significant inconsistencies even when exclusively using the best translations.

\subsection{Consistency and performance}\label{subsec:consistency_performance}

While we have shown that the model's inconsistency does not stem from poor translation quality, it could still stem from an inability to perform the task, leading to somewhat ``random'' responses on different task versions.
To investigate this hypothesis we look at the model's accuracies.

\paragraph{Task accuracies.}
In \cref{tab:performance} (column $T_{src}$), we report the monolingual task accuracies for the model on both tasks, for all languages.
Accuracies are generally higher for PAWS-X (with only two class labels) than XNLI (with three class labels).
In particular, accuracies for German on XNLI are very low.
\cref{app:ablations_external} presents the accuracies for various combinations of input data and instruction languages, which indicate that the model struggles with the German instruction (rather than input) for XNLI.
Furthermore, the accuracies for English are higher than for other languages.
While this may not be surprising given the predominantly English training data, it does raise an intriguing point: if a model can perform a particular task in English, and it can correctly translate the task into a different language, why is it not able to perform the task at a similar level in that other language?

\begin{table}[b]
\centering
\begin{tabular}{l|l|l|c|c}
\multicolumn{3}{c|}{ } & \multicolumn{2}{c}{\textbf{Accuracy}}\\
                            & \textbf{Src} & \textbf{Tgt} & $T_\text{src}$ & $T_{src\rightarrow tgt}$\\
\hline
\multirow{4}{*}{PAWS-X}     & en  & de & \multirow{2}{*}{0.77}    &  0.76 \\
                            & en  & zh  &                         &  0.66 \\
                            \cline{2-4}
                            & de & en  & 0.71                     &  0.73 \\
                            & zh & en  & 0.60                     &  0.68 \\
\hline
\multirow{4}{*}{XNLI}       & en & de  & \multirow{2}{*}{0.71}    &  0.60 \\
                            & en & zh  &                          &  0.60 \\
                            \cline{2-4}
                            & de & en  & 0.48                    &  0.65 \\
                            & zh & en  & 0.56                     &  0.61 \\
\end{tabular} 
\caption{Accuracies on PAWS-X and XNLI for the original task $T_{src}$, and model-internal translations $T_{src\rightarrow tgt}$ from source (src) to target (tgt) language.}
\label{tab:performance}
\end{table}

To further investigate this point, we now consider the accuracies of the model on the task using the model's own translation, which we report in \cref{tab:performance} (column $T_{src\rightarrow tgt}$).
Accuracies for translating either instructions or input sentences only can be found in \cref{app:all_internal_accuracies}.
The results confirm our earlier observation that the model does not maintain consistent meaning representations across languages: even though translations are generated by the model itself and thus should be meaning-preserving according to the model, they lead to differences in performance (compared to the baselines in column $T_{src}$).

These differences in performance also have practical consequences.
While translating from English to German or Chinese leads to a decrease in accuracy, translating from German or Chinese to English leads to an \textit{increase} in accuracy for both PAWS-X and XNLI.
Such improvements can also be observed when translating to English from other languages, like French and Spanish, and are largely due to translating the instruction (see \cref{app:all_internal_accuracies}).
It seems that the model's language-dependent task understanding -- especially interpretation -- can be exploited to increase performance on ``lower''-resource languages by instructing the model to first translate the incoming prompt to English and then to perform the task.

\paragraph{Consistent correct vs incorrect.}
We further investigate if there is a difference in consistency between examples for which the model provides a correct answer and those for which it provides an incorrect answer.
This comparison is interesting because correct and incorrect consistent examples provide different levels of evidence for the consistency of a model.
Being consistently \emph{incorrect} across two examples points to an error in the model's understanding but provides stronger evidence for the consistency of its underlying representations than examples that are consistently correct.
Whereas the latter are correct in both languages and could, in theory, have been inferred independently from the data for those respective languages, it is more unlikely that a model makes an identical but unrelated incorrect inference in two different languages.

First, we establish a baseline, by computing the consistency between two runs with the same $T_{en}$ inputs (\cref{tab:consistency-details}, first column, row 1 for each task, respectively). 
The overall consistencies for this baseline are very high: $99\%$ for PAWS-X and $98\%$ for XNLI.
Accordingly, when asked the same question multiple times, the model usually gives the same response.
In the second and third row (per task, respectively), we further break down consistency and compute what percentage of the correct and incorrect examples were consistent.
As we can see, the baseline case has a high consistency for incorrect responses ($98\%$ and $96\%$), implying that the model's errors are systematic and not due to random guessing. 

Moving to the model-internal translations, we observe a general decrease in consistency that affects both correct and incorrect responses.
However, the consistency for incorrect examples is notably lower than for correct examples.
Given that the model's errors are systematic, this discrepancy suggests that at least some of the consistently correct examples have been inferred independently in both languages.
In conclusion, the comparatively low consistencies for incorrect examples provide corroborating evidence for a sense-dependent task understanding.

\begin{table*}[ht!]
\centering
\begin{tabular}{l|l|ccccc}
\multicolumn{2}{c|}{ }  &  $T_{en}$ & $T_{en\rightarrow de}$ & $T_{en\rightarrow zh}$ & $T_{de\rightarrow en}$ & $T_{zh\rightarrow en}$ \\
\hline
\multirow{3}{*}{PAWS-X}  & consistency all               & 0.99  & 0.84 & 0.76 & 0.86 & 0.70 \\
                         & consistency correct          & 0.99  & 0.89 & 0.78 & 0.92 & 0.82 \\
                         & consistency incorrect        & 0.98  & 0.67 & 0.71 & 0.72 & 0.52 \\
\hline
\multirow{3}{*}{XNLI}    & consistency all               &  0.98   & 0.74 & 0.67 & 0.63 & 0.67 \\
                         & consistency correct          &  0.99 & 0.77 & 0.71 & 0.83 & 0.80 \\
                         & consistency incorrect        &  0.96 & 0.66 & 0.57 & 0.45 & 0.50

\end{tabular}
\caption{Detailed consistencies for the core experiment as well as for a baseline of two different runs with $T_{en}$.
Listed are the consistency across all responses (consistency all), as well as the consistency across responses that were correct (consistency correct) and responses that were incorrect (consistency incorrect) on the source task.}
\label{tab:consistency-details}
\end{table*}

\section{Conclusion}

In this paper, we presented a novel paradigm for evaluating language models, which leverages consistency across different linguistic senses.
Our method can be used to assess generalisation ability beyond specific forms.
It offers affordability and applicability to different evaluation tasks, while also mitigating the risk of evaluating on data that the model has already encountered during training.
As such, multisense evaluation is not an \textit{alternative} to current benchmarks but a \textit{complement}. 
Reporting consistency next to standard evaluation metrics will make model evaluation more meaningful in providing an estimate of how well the model understands a given task beyond its specific form.
Therefore, we encourage other researchers to treat multisense consistency as an essential part of benchmarking. 

To showcase the effectiveness of our paradigm, we conducted a \textit{multilingual} multisense consistency evaluation of ChatGPT (gpt-3.5-turbo), a SOTA LLM.
The results of this evaluation unveiled significant inconsistencies across different senses generated by the model itself through translation, suggesting a lack of genuine, cross-sense task understanding.
To ensure the validity of this interpretation, we ruled out alternative explanations such as model-subjective or objective changes in meaning caused by the translation as well as inadequate performance on the original task.
Collectively, these findings show that ChatGPT exhibits a language- and therefore sense-dependent task understanding, which might also affect other leading LLMs.

Our paradigm can be cheaply and easily expanded to include more languages, tasks, models, and notions of ``sense''.
Our choice to generate multiple senses through translation is well-suited for evaluating current and future models, given the growing trend towards multilingual models with increasingly proficient translation abilities.
Nevertheless, numerous other multisense evaluations are conceivable.
For instance, instead of using model-internal translations, one could employ model-internal paraphrases.
Multiple senses could also be generated in different domains, such as arithmetic (different formulas yielding the same result) or code (different functions producing the same input-output mapping). 
Last but not least, calculating consistency for various tasks may help disentangle ``unfounded'' language-specific differences (forming the focus of our analysis) from differences related to cultural bias.

In conclusion, multisense consistency can be applied as long as the model under investigation can create different senses of a given task and has some understanding of the task in its original sense.
It offers the possibility of evaluating the task understanding detached from a specific task realisation, and we hope it will contribute to making standard benchmark evaluations more meaningful.

\section*{Limitations}

While our method can certainly be extended to other tasks and models, some of these extensions may prove more challenging than others.
In particular, evaluating consistency between model responses that are more variable than the ones in our experiments is less straightforward, and requires an appropriate definition of consistency.
More variable responses may arise when working with LLMs that have not been adapted to deal with instructions.
For example, we instruct ChatGPT to choose a response from a set of predefined responses (\textit{Please answer with ``yes'' or ``no''}, \textit{Please answer with ``A'', ``B'', or ``C''}) and it largely follows these instructions.
A standard LLM may deviate from these answer templates, leading to complications in calculating the consistency.
In addition, more variable responses may arise when dealing with tasks that do not correspond to a classification problem.
Even testing factual knowledge with a question-answering task may lead to variations in responses.
For example, model responses like \textit{5}, \textit{5 times}, or \textit{five}, may be consistent but are different.
Further, the model can generate responses that are only partially overlapping, e.g. \textit{disastrous financial situation} versus \textit{bad financial situation}, which might require a graded definition of consistency.
Thus, moving forward, it is important to develop appropriate definitions of consistency as well as corresponding automatic evaluation procedures.
Given that judging whether two answers have the same meaning is much easier than providing these answers, the consistency evaluation might even be outsourced to the model under investigation.

\section*{Ethics statement}

We proposed a novel method for evaluating the self-consistency of LLMs by using the models themselves to generate alternative forms or senses of the same task.
If a model is self-consistent according to this evaluation, its task understanding goes beyond matching patterns that are present in specific forms.
Importantly, though, the model can still be subject to the many problems that currently pertain to pretrained LLMs such as hallucinations or biases.
Thus, when using the model to generate different forms or to make predictions for a certain task, its output may contain wrong information, as well as biased and offensive content.
These problematic outputs may or may not lead to inconsistencies, and as discussed in the conclusion, future work could try to employ multisense consistency as a tool to detect them.
As of now, however, multisense consistency is a means to evaluate a model's robustness, not a means to determine whether the content of its answers is desirable.

\section*{Acknowledgements}

We would like to thank Felipe Cerdas, Xin Huang, and Vera Lamprecht for their help with the translations.
In addition, we would like to thank Marco Baroni and Ryan Nefdt for giving us feedback on an earlier version the draft.

\bibliography{anthology,custom}
\bibliographystyle{acl_natbib}

\appendix

\section{Genbench evaluation card}\label{app:genbench}

Our work uses generalisation across senses to assess task understanding in LLMs.
In \cref{fig:genbench_eval_card}, we provide the GenBench eval card \citep{hupkes2023stateoftheart} of our experiments.

\begin{figure}[H]
    \centering
\newcommand{\tabularwidth}{\columnwidth}

\newcommand{\expone}{$\boxtimes$}

\renewcommand{\arraystretch}{1.1}
\setlength{\tabcolsep}{0mm}
\begin{tabular}{|p{\tabularwidth}<{\centering}|}
\hline

\rowcolor{gray!60}
\textbf{Motivation} \\
\footnotesize
\begin{tabular}{p{0.25\tabularwidth}<{\centering} p{0.25\tabularwidth}<{\centering} p{0.25\tabularwidth}<{\centering} p{0.25\tabularwidth}<{\centering}}
\textit{Practical} & \textit{Cognitive} & \textit{Intrinsic} & \textit{Fairness}\\
& 		
& \expone 
& 		

\end{tabular}\\

\rowcolor{gray!60}
\textbf{Generalisation type} \\
\footnotesize
\begin{tabular}{p{0.17\tabularwidth}<{\centering} p{0.20\tabularwidth}<{\centering} p{0.14\tabularwidth}<{\centering} p{0.17\tabularwidth}<{\centering} p{0.18\tabularwidth}<{\centering} p{0.14\tabularwidth}<{\centering}}
\textit{Compositional} & \textit{Structural} & \textit{Cross Task} & \textit{Cross Language} & \textit{Cross Domain} & \textit{Robust- ness}\\
& 		
& 		
& \expone		
& 		
& \hspace{0.2mm} \expone		
\end{tabular}\\

\rowcolor{gray!60}
\textbf{Shift type} \\
\footnotesize
\begin{tabular}{p{0.25\tabularwidth}<{\centering} p{0.25\tabularwidth}<{\centering} p{0.25\tabularwidth}<{\centering} p{0.25\tabularwidth}<{\centering}}
\textit{Covariate} & \textit{Label} & \textit{Full} & \textit{Assumed}\\
\expone\hspace{0.8mm}		
& 		
& 		
& 		

\vspace{2mm} \\
\end{tabular}\\

\rowcolor{gray!60}
\textbf{Shift source} \\
\footnotesize
\begin{tabular}{p{0.25\tabularwidth}<{\centering} p{0.25\tabularwidth}<{\centering} p{0.25\tabularwidth}<{\centering} p{0.25\tabularwidth}<{\centering}}
\textit{Naturally occuring} & \textit{Partitioned natural} & \textit{Generated shift} & \textit{Fully generated}\\
\expone\hspace{0.8mm}		
& 		
& 		
& 		
\vspace{2mm} \\
\end{tabular}\\

\rowcolor{gray!60}
\textbf{Shift locus}\\
\footnotesize
\begin{tabular}{m{0.22\tabularwidth}<{\centering} p{0.28\tabularwidth}<{\centering} p{0.24\tabularwidth}<{\centering} m{0.24\tabularwidth}<{\centering}}
\textit{Train--test} & \textit{Finetune train--test} & \textit{Pretrain-- train} & \textit{Pretrain-- test}\\
& 		
& 		
& \hspace{3.3mm}\expone		
\end{tabular}\\
\hline
\end{tabular}

    \caption{Our experiments assess cross-lingual generalisation for natural corpora, in pretrained LLMs, to assess LLM task understanding.}
    \label{fig:genbench_eval_card}
\end{figure}

\section{Task instructions}\label{app:instructions}

Table \ref{tab:instructions} shows the task instructions for both tasks, in all languages.
The table shows the original English, German, and Chinese instructions, as well as the model-internal translations of these instructions.\footnote{Note that we also accept \begin{CJK*}{UTF8}{gbsn}不是\end{CJK*} instead of \begin{CJK*}{UTF8}{gbsn}否\end{CJK*} for $I_{zh}$.}

\begin{table*}

\centering
\label{crouch}

\begin{tabular}{l|l| p{11cm} }
\hline
\hline 
\textbf{Task}      
& \textbf{Language}   
& \textbf{Instruction} \\
\hline

\multirow{13}{*}{{PAWS-X}} &

$I_{en}$ & 
\small \textit{Do the following sentences have the same meaning? Sentence 1: ``[sentence\_1]'' Sentence 2: ``[sentence\_2]'' Please answer with ``yes'' or ``no''.}
\\
\cline{2-3}

& $I_{de}$ &
\small \textit{Haben die folgenden Sätze die gleiche Bedeutung? Satz 1: ``[sentence\_1]'' Satz 2: ``[sentence\_2]'' Bitte antworte mit ``ja'' oder ``nein''.}
\\
\cline{2-3}

& $I_{zh}$ & 
\begin{CJK*}{UTF8}{gbsn}
        \small \textit{下面的句子有着相同的含义吗？ 句子 1: ``[sentence\_1]'' 句子 2: ``[sentence\_2]'' 请用“是”或者“否”回答。}
\end{CJK*}
\\
\cline{2-3}

& $I_{en\rightarrow de}$ & 
\small \textit{Haben die folgenden Sätze die gleiche Bedeutung? Satz 1: ``[sentence\_1]'' Satz 2: ``[sentence\_2]'' Bitte antworten Sie mit ``Ja'' oder ``Nein''.}
\\
\cline{2-3}

& $I_{en\rightarrow zh}$ & 
\begin{CJK*}{UTF8}{gbsn}
        \small \textit{以下句子的意思相同吗？句子 1: ``[sentence\_1]'' 句子 2: ``[sentence\_2]'' 请用“是”或“不是”回答。}
\end{CJK*}
\\
\cline{2-3}

& $I_{de\rightarrow en}$ & 
\small \textit{Do the following sentences have the same meaning? Sentence 1: ``[sentence\_1]'' Sentence 2: ``[sentence\_2]'' Please respond with `yes' or `no'.}
\\
\cline{2-3}

& $I_{zh\rightarrow en}$ &
\small \textit{Does the following sentence have the same meaning? Sentence 1: ``[sentence\_1]'' Sentence 2: ``[sentence\_2]'' Please answer with `yes' or `no'.}
\\
\hline

\multirow{21}{*}{{XNLI}} &

$I_{en}$ & 
\small \textit{Given the following sentence pair, which one of the following is true: (A) the first sentence entails the second sentence, (B) the first sentence contradicts the second sentence, or (C) neither of the two? Sentence 1: ``[sentence\_1]'' Sentence 2: ``[sentence\_2]'' Please answer with ``A'', ``B'', or ``C''.}
\\
\cline{2-3}

& $I_{de}$ &
\small \textit{Welche dieser Aussagen trifft auf das folgende Satzpaar zu: (A) der erste Satz impliziert den zweiten Satz, (B) der erste Satz widerspricht dem zweiten Satz, oder (C) keines von beiden? Satz 1: ``[sentence\_1]'' Satz 2: ``[sentence\_2]'' Bitte antworte mit ``A'', ``B'' oder ``C''.}
\\
\cline{2-3}

& $I_{zh}$ & 
\begin{CJK*}{UTF8}{gbsn}
        \small \textit{对于给出的一对句子，以下哪一个选项是正确的：（A）第一个句子涵盖了第二个句子（B）第一个句子与第二个句子矛盾 （C）两者都不？句子 1: ``[sentence\_1]'' 句子 2: ``[sentence\_2]'' 请用“A”、“B”或“C”来回答。}
\end{CJK*}
\\
\cline{2-3}

& $I_{en\rightarrow de}$ & 
\small \textit{Angesichts des folgenden Satzpaares, welche der folgenden Aussagen ist wahr: (A) Der erste Satz impliziert den zweiten Satz, (B) Der erste Satz widerspricht dem zweiten Satz oder (C) Keines von beiden? Satz 1: ``[sentence\_1]'' Satz 2: ``[sentence\_2]'' Bitte antworten Sie mit ``A'', ``B'' oder ``C''.}
\\
\cline{2-3}

& $I_{en\rightarrow zh}$ & 
\begin{CJK*}{UTF8}{gbsn}
        \small \textit{给定以下句子对，哪一个是正确的：（A）第一句蕴含第二句，（B）第一句与第二句相矛盾，还是（C）两者都不是？句子 1: ``[sentence\_1]'' 句子 2: ``[sentence\_2]'' 请用``A''、``B''或``C''回答。}
\end{CJK*}
\\
\cline{2-3}

& $I_{de\rightarrow en}$ & 
\small \textit{Which of these statements applies to the following pair of sentences: (A) the first sentence implies the second sentence, (B) the first sentence contradicts the second sentence, or (C) neither of the above? Sentence 1: ``[sentence\_1]'' Sentence 2: ``[sentence\_2]'' Please reply with ``A'', ``B'', or ``C''.}
\\
\cline{2-3}

& $I_{zh\rightarrow en}$ &
\small \textit{For a given pair of sentences, which of the following options is correct: (A) The first sentence covers the second sentence. (B) The first sentence contradicts the second sentence. (C) Neither of them? Sentence 1: ``[sentence\_1]'' Sentence 2: ``[sentence\_2]'' Please answer with ``A'', ``B'', or ``C''.}\\
\hline
\end{tabular}
\caption{Task instructions in different languages. The original instructions in English, German, and Chinese are given by $I_{en}$, $I_{de}$, and $I_{zh}$. The model-internal translations of these instructions (from source to target language) are given by $I_{source\rightarrow target}$.}
\label{tab:instructions}
\end{table*}

\section{Translation instructions}\label{app:translation_instructions}

We used the following instructions for model-internal translations:
\begin{itemize}
    \item en$\rightarrow$de/zh:\\ \textit{ Please translate the following text into \\German/Chinese: ``[text]''}
    \item de$\rightarrow$en:\\\textit{ Bitte übersetze den folgenden Text ins \\Englische: ``[text]''}
    \item zh$\rightarrow$en:\\
\textit{\begin{CJK*}{UTF8}{gbsn} 请将下面的文字翻译成英语\end{CJK*}: ``[text]''}

\end{itemize}

\section{Elaborations on the inconsistencies in the main experiment}\label{app:elaboration_main_experiment}

Tables \ref{tab:label_distributions_paws} and \ref{tab:label_distributions_xnli} display the distributions of the ground truth labels and the predicted labels for different representations of PAWS-X and XNLI, respectively.
Regarding PAWS-X (see \cref{tab:label_distributions_paws}), the model consistently overestimates the number of paraphrases across all task representations.
At the same time, the predicted label distributions vary -- sometimes strongly -- between the original task versions ($T_{en}$, $T_{de}$, $T_{zh}$) and their model-internal translations.
For example, the amount of predicted paraphrases increases from $58$\% in English to $62$\% when translating to German and $68$\% when translating to Chinese.
More extremely, the model predicts $78$\% paraphrases on the Chinese task version but only $53$\% on its translation to English.
These distributions suggest the presence of language-dependent biases in the model's assessment of whether two sentences convey the same meaning or not.
In particular, if the model translates from a certain source language to a certain target language, the predicted label distribution for the model-internal translation ($T_{source\rightarrow target}$) becomes more similar to that of the ``model-external'' translation ($T_{target}$).
In other words, if the model predicts more or fewer paraphrases on the target language ($T_{target}$) compared to the source language ($T_{source}$), the predictions on the model-internal translation tend to increase or decrease accordingly.

\begin{table}
\centering
\begin{tabular}{l| c c}
    & \multicolumn{2}{c}{\textbf{label}} \\
    &   true & false \\
    \hline
    ground truth & 0.45 & 0.55 \\
    \hline
    $T_{en}$ & 0.58 & 0.42 \\
    $T_{en\rightarrow de}$ & 0.62 & 0.38 \\
    $T_{en\rightarrow zh}$ & 0.69 & 0.31 \\
    \hline
    $T_{de}$ & 0.65 & 0.34 \\
    $T_{de\rightarrow en}$ & 0.62 & 0.38 \\
    \hline
    $T_{zh}$ & 0.78 & 0.21\\
    $T_{zh\rightarrow en}$ & 0.53 &  0.47 \\
\end{tabular}
\caption{Ground truth and predicted label distributions for PAWS-X.}
\label{tab:label_distributions_paws}
\end{table}

These patterns are reflected in the types of inconsistencies observed when comparing the model's responses on the original task version to those on the model-internal translation.
When translating from English to German, $60\%$ of the inconsistencies are cases where the model classifies a sentence pair as a paraphrase in German but not in English.
When translating from English to Chinese (with an even higher proportion of predicted paraphrases), these cases account for $0.72$\% of the inconsistencies.
Conversely, when translating from German or Chinese to English, most inconsistencies are cases where the model classifies a sentence pair as a paraphrase in the source language but not in English ($60\%$ for German and $92$\% for Chinese).

For XNLI (see \cref{tab:label_distributions_xnli}), the model consistently overestimates the number of entailments and, correspondingly, tends to underestimate the number of contradicting and neutral sentence pairs.
Especially notable are the high amounts of predicted entailments for $T_{de}$ ($69$\%) and $T_{en\rightarrow de}$ ($62$\%), which are further explored in the qualitative analysis provided in \cref{app:qualitative_analysis}.
Despite this general trend, the predicted distributions exhibit significant variations between the source language and the model-internal translation.
For example, while the model predicts only $48$\% entailments on $T_{en}$, it predicts $62$\% on $T_{en\rightarrow de}$.
Conversely, while it predicts $69$\% entailments on $T_{de}$, it predicts only $54$\% on $T_{de\rightarrow en}$.

\begin{table}
\centering
\begin{tabular}{l| c c c}
    & \multicolumn{3}{c}{\textbf{label}} \\
    &   entail & neutral & contradict \\
    \hline
    ground truth & 0.33 & 0.33 & 0.33 \\
    \hline
    $T_{en}$ & 0.48 & 0.21 & 0.30 \\
    $T_{en\rightarrow de}$ & 0.62 & 0.18 & 0.20 \\
    $T_{en\rightarrow zh}$ & 0.42 & 0.36 & 0.22 \\
    \hline
    $T_{de}$ & 0.69 & 0.26 & 0.05 \\
    $T_{de\rightarrow en}$ & 0.54 & 0.15 & 0.31 \\
    \hline
    $T_{zh}$ & 0.52 & 0.25 & 0.22 \\
    $T_{zh\rightarrow en}$ & 0.40 &  0.31 & 0.30 \\
\end{tabular}
\caption{Ground truth and predicted label distributions for XNLI.}
\label{tab:label_distributions_xnli}
\end{table}

Compared to PAWS-X, it is more challenging to identify patterns in the inconsistencies for XNLI.
Firstly, there are more interactions between inconsistencies as there are three class labels instead of two.
Secondly, the more complex task instruction may have a stronger influence, leading to mixed effects from differences in task interpretation and execution.
However, for translations between English and German (which are also of higher quality than translations between English and Chinese), some patterns can still be identified.
Most inconsistencies when translating from English to German involve sentences where the model switches from \textit{neutral} ($33$\%) or \textit{contradiction} ($35$\%) to \textit{entailment}, together accounting for $68$\% of the inconsistencies.
When translating from German to English, in turn, a large proportion of the sentence pairs formerly classified as \textit{entailment} are now classified as \textit{contradiction}, constituting $51$\% of the inconsistencies (with $9$\% for switching from \textit{entailment} to \textit{neutral}).
These inconsistencies might be considered a particularly strong argument against a genuine task understanding by the model, as it regularly switches interpretation between the contrasting concepts of ``entailment'' and ``contradiction'', rather than mostly transitioning between \textit{neutral} and the other two categories.

\section{Qualitative analysis for model-internal translations from English to German}\label{app:qualitative_analysis}

We conduct a qualitative analysis of the model's inconsistencies when translating from English to German.
We examine 100\% of the inconsistencies on PAWS-X (329 data points) and the first 50\% of the inconsistencies on XNLI (664 data points).

To begin with, our focus lies on verifying whether the model's change in response is indeed due to a change in sense (but not meaning) or whether there might be an alternative explanation.
For that purpose, we classify the data points into two categories: category (1) if no alternative explanation can be identified and category (2) if an alternative explanation can be identified.
After reviewing the examples, we define the following alternative explanations for category (2):

\paragraph{(2.1) Ambiguity}
    \begin{itemize}
        \item Source ambiguities: the source sentence contains ambiguous expressions and the model a) switches interpretation or b) resolves this ambiguity.
        \item Target ambiguities: the target sentence contains ambiguous expressions that were not ambiguous in the source sentence.
    \end{itemize}

\paragraph{(2.2) Translation quality}
    \begin{itemize}
        \item The translation does not preserve meaning.
        \item The translation is of poor linguistic quality, potentially making the task more difficult.
    \end{itemize}

\paragraph{(2.3) Identical sentences}
    \begin{itemize}
        \item The translations of the input sentences are identical, which confuses the model.
    \end{itemize}

Note that this is a very conservative encoding.
Firstly, some of these cases should arguably not cause inconsistencies.
For example, if the model  ``understands'' what it means for two sentences to have the same meaning, it should also understand that two identical sentences have the same meaning (subcategory 2.3).
Secondly, even if there is ambiguity in the source or target language, or the linguistic quality is subpar, it is not clear whether the model changes its response because of these factors.

Here are examples illustrating the subcategories.
An example of ambiguity is the following sentence pair from PAWS-X:
``\textit{The film stars Oscar Nunez, Rob Huebel, Timothée Chalamet, Lily Rabe, Anthony Quintal, and Lili Reinhart.}'',
``\textit{Film stars Oscar Nunez, Rob Huebel, Timothée Chalamet, Lily Rabe, Anthony Quintal, and Lili Reinhart.}''
The first sentence is ambiguous as to whether \textit{stars} is a verb or part of the compound noun \textit{film stars}.
In German, it is translated as a verb (``\textit{Der Film hat Oscar Nunez, [...]}''), and as a result the sentence pair is classified as a paraphrase in English but not in German.
An example of an inaccurate translation is the following sentence pair from XNLI:
``\textit{Smaller boats for local jaunts can be rented at Sea Horse Boat Rentals, Marsh Harbour, Abacos (Tel.,}'',
``\textit{You can rent one passenger boats.}''
Due to the missing hyphen between \textit{one} and \textit{passenger} in the premise, \textit{one passenger boats} is interpreted as \textit{one passenger boat} in the German translation (``\textit{Sie können ein Passagierboot mieten.}'').
The model correctly predicts that the sentences are \textit{neutral} in German but predicts \textit{entailment} in English.
Finally, here is an example of identical sentences from PAWS-X:
``\textit{The first series was recorded by critics better than the second .}'',
``\textit{The first series was better received by critics than the second .}'';
which are both accurately translated to ``\textit{Die erste Staffel wurde von Kritikern besser aufgenommen als die zweite.}''
While the model predicts that the sentences have the same meaning in English, it only replies that the sentences are identical in German (``\textit{Die beiden Sätze sind identisch}.'').

Despite the conservative encoding, a majority of the inconsistencies -- $78$\% for PAWS-X and $86$\% for XNLI -- fall into category (1), which means that none of the alternative explanations are applicable.
For PAWS-X, $6$\% of the inconsistencies may be related to ambiguities, $10$\% to translation quality, and $4$\% to identical sentence pairs.\footnote{The remaining $2$\% are sentences that do not fall into category (1) or (2) because the original sentences are so ungrammatical that it is difficult to determine whether the translation is accurate.}
For XNLI, it is $7$\% for ambiguities, $7$\% for translation quality, and $0$\% for identical sentence pairs.
Alongside the analyses in \cref{subsec:translation_evaluation} and \cref{subsec:consistency_performance}, this less general but more in-depth analysis provides further evidence that the model's responses are sense-dependent.

Examining examples from category (1) can help understand how a sense-dependent task understanding might lead to inconsistencies.
In most cases, it remains unclear why the model makes different predictions.
Especially for PAWS-X, it is surprising how the model is sometimes fooled by the adversarial nature of the sentences in one language but not the other.
For example, given the sentence pair ``\textit{The Tabaci River is a tributary of the River Leurda in Romania .}'', ``\textit{The Leurda River is a tributary of the Tabaci River in Romania .}'';
and the correct German translations ``\textit{Der Fluss Tabaci ist ein Nebenfluss des Flusses Leurda in Rumänien.}'', ``\textit{Der Fluss Leurda ist ein Nebenfluss des Tabaci-Flusses in Rumänien.}'';
the model identifies that the sentences have different meanings in English but not in German.
The reverse case where the model is fooled in English but not in German also exists.

In some cases, one can speculate that certain informational content of the sentences is more readily available to the model in one language than the other, which might influence its response.
Take for example the following sentence pair from XNLI:
``\textit{Among the many jazz clubs are the famed Jazz Bakery in Culver City, the Catalina Bar and Grill in Hollywood, and the Baked Potato in North Hollywood.}'',
``\textit{There are no famous jazz clubs in Los Angeles.}''
The model correctly predicts \textit{contradiction} in English but predicts \textit{entailment} in German, possibly because the information that Culver City is part of Los Angeles appears more often in English text than in German text.
This example constitutes an important exception because, unlike for most other examples, the ability to make a correct prediction is knowledge-dependent.
As such, it illustrates the very situation where the model should give the same response in both languages.
The fact that the model apparently knows that Culver City is in LA when asked in English but not when asked in German thus provides powerful evidence for a sense-dependent task understanding.

An influence of prior knowledge may also lead to mistakes as in the following example from PAWS-X:
``\textit{Stipsits was born in Korneuburg , Germany and spent his childhood in Stammersdorf , Vienna .}'',
``\textit{Stipsits was born in Korneuburg , and spent his childhood in Stammersdorf , Vienna .}''
The model correctly classifies these as paraphrases in English, but argues that the sentences do not have the same meaning in German because Korneuburg is in Austria and not in Germany (``\textit{Nein. Satz 1 ist inkorrekt, da Korneuburg in Österreich liegt und nicht in Deutschland [...]}'').
It has very recently been established that LLMs (including ChatGPT) use prior knowledge for language inference, for example, they judge the truth of the hypothesis based on information in the training data rather than information in the premise~\citep{mckenna2023sources}.
Our results are in line with this finding and further suggest that the interfering information is language-dependent.

Other cases where the model provides an explanation for its answer (deviating from the answer template) are also revealing.
In particular, they show that the model generally struggles to interpret the German instruction for XNLI, consistent with the low accuracies for $T_{en\rightarrow de}$ (0.60) and $T_{de}$ (0.48) in \cref{tab:performance}.
For example, on one occasion the model responds ``\textit{Die richtige Antwort ist (C) Keines von beiden. Die beiden Sätze sind unabhängig voneinander und widersprechen sich nicht.}'', on another one
``\textit{Die richtige Antwort ist (C) Keines von beiden. Die beiden Sätze haben keine direkte Beziehung zueinander und widersprechen sich auch nicht.}''
These responses indicate that the model excludes the option of entailment because the sentences \textit{are independent from each other} or \textit{do not have a direct relationship}.
Possibly, the model also applies these as positive criteria for entailment, which would explain why it significantly overestimates the number of entailments in German.

\section{Main experiment with BoolQ}\label{app:boolq}

BoolQ is a question answering dataset where each example consists of a passage and a yes/no question about that passage.
We use the validation split of the dataset and prompt the model by providing the passage, followed by the question (capitalised and with a question mark), and the instruction \textit{Please answer with ``yes'' or ``no''.}
We retrieve the model's responses for $T_{en}$ and evaluate consistency with $T_{en\rightarrow de}$ and $T_{en\rightarrow zh}$, respectively.
The translations of the input sentences are obtained by instructing the model to translate each passage and question in a separate request, using the English translation instruction (see \cref{app:translation_instructions}).
Since the instruction for BoolQ corresponds to the instruction suffix for PAWS-X, we reuse these translations.
The resulting consistencies are provided in \cref{tab:results_boolq}, together with the accuracy for each task version.
The consistencies follow the same pattern as those for PAWS-X and XNLI when translating from English to German and Chinese (see \cref{tab:consistency}):
The model is not perfectly consistent regardless of the target language, with lower consistency for the Chinese translation.

\begin{table}
\centering
\begin{tabular}{l|ccc}
\textbf{Task version} &\textbf{Consistency} & \textbf{Accuracy} \\
\hline
$T_{en}$ (orig)          &  -     &  0.86  \\
$T_{en\rightarrow de}$   &  0.89  &  0.82  \\
$T_{en\rightarrow zh}$   &  0.81  &  0.78
\end{tabular}
\caption{Consistency and accuracy for BoolQ.
The first column provides the consistencies between the model's responses on the original task ($T_{en}$) and the model-internal translations of that task to German ($T_{en\rightarrow de}$) and Chinese ($T_{en\rightarrow zh}$).
The second column provides the model's accuracy for each task version.
}
\label{tab:results_boolq}
\end{table}

\section{Translation evaluation scores}\label{app:translation_evaluation_scores}

We evaluate translation quality for the input sentences using BLEU, ROUGE, and COMET-22 scores (see \cref{tab:translation_scores}).

\begin{table*}
\centering
\begin{tabular}{l|l|ccccc}
 &   \textbf{Src}$\rightarrow$\textbf{Tgt} &  \textbf{BLEU} &  \textbf{Rouge1} &  \textbf{Rouge2} &  \textbf{Rouge-l} &  \textbf{COMET-22} \\
\hline
\multirow{4}{*}{PAWS-X} &  en$\rightarrow$de    & 56.5 &   0.80 &   0.64 &    0.77 &  0.89 \\
                        & en$\rightarrow$zh     & 49.2 &   0.68 &   0.42 &    0.62 &  0.86 \\
                        & de$\rightarrow$en     & 60.0 &   0.87 &   0.72 &    0.83 &  0.88 \\
                        & zh$\rightarrow$en     & 37.6 &   0.73 &   0.49 &    0.66 &  0.85 \\
\hline
\multirow{4}{*}{XNLI}  &  en$\rightarrow$de     & 41.4 &   0.71 &   0.52 &    0.68 &  0.88 \\
                        & en$\rightarrow$zh     & 43.5 &   0.66 &   0.39 &    0.62 &  0.87 \\
                        & de$\rightarrow$en     & 45.8 &   0.76 &   0.57 &    0.74 &  0.89 \\
                        & zh$\rightarrow$en     & 28.0 &   0.61 &   0.37 &    0.57 &  0.86 \\
\end{tabular}
\caption{Evaluation of the model-internal translation of the input data.}
\label{tab:translation_scores}
\end{table*}

\section{Inconsistencies for very high quality translations} \label{app:translation_quality-consistency}

We extend the analyses from Section \ref{subsec:translation_evaluation} by calculating the inconsistencies for data points with a BLEU score of at least $50$. Our focus remains on translations of the input data and the model is instructed in the original (source) language. Table \ref{tab:consistency-thresholded} shows the amount of data (\%) included in the analysis, along with the corresponding consistency. Importantly, the model's inconsistency persists, as it never achieves consistencies surpassing $0.87$. Moreover, across the board, consistencies exhibit only a slight improvement compared to the original values (see \textit{consistency orig}, same as in Table \ref{tab:consistency}, column \textit{X}). The only substantial increase in performance occurs for translations from Chinese to English on XNLI, with consistency rising from $0.72$ to $0.80$. This finding aligns with the observation that the translations from Chinese to English are of significantly lower quality than the other translations. Hence, bad translations may reduce consistency, but this phenomenon is only observed in one specific case. 

\begin{table*}
\centering
\begin{tabular}{l|l|cccc}
 & & en$\rightarrow$de & en$\rightarrow$zh & de$\rightarrow$en & zh$\rightarrow$en \\
\hline
\multirow{3}{*}{PAWS-X} & consistency orig  & 0.85 & 0.79 & 0.86 & 0.75 \\
                        \cline{2-6}
                        & consistency BLEU $>50$   & 0.86     &   0.82   &    0.87  &   0.78\\
                        & \% included BLEU $>50$  & 56.6      &   40.1    &    67.1   &   20.6 \\
\hline
\multirow{3}{*}{XNLI}   & consistency (orig)  & 0.76 & 0.71 & 0.81 & 0.72 \\
                        \cline{2-6}
                        & consistency BLEU $>50$  & 0.77     &   0.72  &   0.82  &  0.80 \\
                        & \% included BLEU $>50$  & 35.6      &   32.3    &   39.6    &  10.5
\end{tabular}
\caption{Only datapoints with BLEU scores of $>50$ are included in this analysis. The table shows the percentage of included data points (\textit{\% included BLEU>50}), and the consistency of the model for these selected translations (\textit{consistency BLEU>50}) compared to the original consistency (\textit{consistency orig}) repeated from Table \ref{tab:consistency} (column \textit{X}).}
\label{tab:consistency-thresholded}
\end{table*}

\section{Performance on mixed languages for input data and instructions}\label{app:ablations_external}

We look at different ablations to understand the effect of using a language other than English for input data or instruction.
\cref{tab:baseline_performance_mixed} shows the model's accuracy on different combinations of languages for input sentences and instructions, always using the input sentences provided by the multilingual benchmark, and the English, German, or Chinese instructions developed for us by native speakers.
Compared to $T_{en}$, with an accuracy of $0.77$ on PAWS-X and $0.71$ on XNLI (see \cref{tab:performance}), accuracy decreases when instruction or input data are changed from English to German or Chinese.
Changing the language for both at the same time further decreases accuracy, as errors from each language change accumulate (see $T_{de}$ and $T_{zh}$ in \cref{tab:performance}).
For PAWS-X there is a more substantial decrease when changing instructions or input data to Chinese compared to German.
For XNLI, especially the use of the German instruction is detrimental, with accuracies dropping from $0.71$ to $0.50$.
Testing alternative German instructions reveals that this effect does not only pertain to our specific formulation.
While a decrease in performance may be expected for non-English inputs, the extent of this effect when changing only the task instruction is surprising.
For example, changing the instruction for PAWS-X from English to Chinese leads to a $10\%$ absolute decrease in accuracy, even though this instruction is very simple.

\begin{table}
\centering
\begin{tabular}{l| c c c c}
& \multicolumn{4}{c}{\textbf{X / I}} \\
& en/de & en/zh & de/en & zh/en \\
\hline
PAWS-X      & 0.75 & 0.67 & 0.73 & 0.68\\
XNLI        & 0.50 & 0.60 & 0.65 & 0.59\\
\end{tabular}
\caption{Accuracies on mixed-language combinations of original input data (\textit{X}) and instructions (\textit{I}).}
\label{tab:baseline_performance_mixed}
\end{table}

\section{Task performance for model-internal translations} \label{app:all_internal_accuracies}

\cref{tab:mixed_performance} shows the model's accuracies for all source languages ($T_{src}$) and the corresponding model-internal translations:
instruction only ($I_{src\rightarrow tgt}$ / $X_{src}$), input sentences only ($I_{src}$ / $X_{src\rightarrow tgt}$), or both ($T_{src\rightarrow tgt}$).
In addition, we add accuracies for French and Spanish and their translations to English.
\cref{subsec:consistency_performance} shows that model-internal translations from German and Chinese to English increase the model's accuracy compared to the original $T_{de}$ and $T_{zh}$ tasks.
The results for French and Spanish show that translations from other languages to English can also increase accuracy.
For instance, translating (input sentences and instruction) from Spanish to English raises the accuracy on PAWS-X from $0.72$ to $0.73$, and on XNLI from $0.60$ to $0.65$.
Looking at the separate effects of translating the instructions or the input sentences to English suggests that the observed improvements can largely be ascribed to the translation of the instruction, regardless of the source language.

\begin{table*}
\centering
\begin{tabular}{l|l|l| c | ccc}
    \multicolumn{3}{c|}{ } & \textbf{Acc (orig)} & \multicolumn{3}{c}{\textbf{Acc (translation)}}\\
    & \textbf{Src} & \textbf{Tgt} & $T_{src}$ & $T_{src\rightarrow tgt}$ &  $I_{src\rightarrow tgt}$ / $X_{src}$ &  $I_{src}$ / $X_{src\rightarrow tgt}$ \\
\hline
\multirow{6}{*}{PAWS-X}     & en & de   & \multirow{2}{*}{0.77} &  0.76  & 0.77   &  0.77 \\
                            & en & zh   &  &  0.66  & 0.75   &  0.70 \\
                            \cline{2-7}
                            & de & en   &  0.71 &  0.73  & 0.72  &  0.70 \\
                            & zh & en   &  0.60 &  0.68  & 0.67  &  0.63 \\
                            \cline{2-7}
                            & fr & en   &  0.72 &  0.72  & 0.72  &  0.71 \\
                            & es & en   &  0.72 &  0.73  & 0.73  &  0.71 \\
\hline
\multirow{6}{*}{XNLI}       & en & de   & \multirow{2}{*}{0.71} &  0.60  & 0.63   &  0.67  \\
                            & en & zh   &  &  0.60  & 0.63   &  0.62  \\
                            \cline{2-7}
                            & de & en   &  0.48 &  0.65  & 0.64   &  0.49 \\
                            & zh & en   &  0.56 &  0.61  & 0.59   &  0.56 \\
                            \cline{2-7}
                            & fr & en   &  0.58 &  0.63  & 0.61   &  0.60 \\
                            & es & en   &  0.60 &  0.65  & 0.67   &  0.60 \\
\end{tabular}
\caption{Accuracies on the original multilingual benchmark tasks ($T_{src}$) and the model-internal translations of these tasks from source (src) to target (tgt) language.
We consider translations of both input data and instructions ($T_{src\rightarrow tgt}$), instruction only ($I_{src\rightarrow tgt}$ / $X_{src}$), and input data only ($I_{src}$ / $X_{src\rightarrow tgt}$).
Besides, we add translations from French and Spanish to English to further study whether translating to English can improve performance.}
\label{tab:mixed_performance}
\end{table*}

\end{document}